\documentclass[journal]{IEEEtran}
\usepackage{cite}
\usepackage{amsmath,amssymb,amsfonts}
\usepackage{algorithmic}
\usepackage{graphicx}
\usepackage{multirow}
\usepackage{textcomp}
\usepackage{xcolor}
\usepackage{inputenc}  
\usepackage[OT1]{fontenc}

\begin{document}
%
\title{Enhancing Remote Sensing Image Retrieval with Triplet Deep Metric Learning Network}
%
%
%

\author{Rui~Cao,
Qian~Zhang,
Jiasong~Zhu,
Qing~Li,
Qingquan~Li,
Bozhi~Liu,
and Guoping~Qiu
\thanks{R. Cao and Q. Zhang are with the International Doctoral Innovation Centre \& School of Computer Science, University of Nottingham Ningbo China, Ningbo 315100, China (e-mail: rui.cao@nottingham.edu.cn).}
\thanks{J. Zhu, Qing Li, and Qingquan Li are with the Shenzhen Key Laboratory of Spatial Smart Sensing and Services \& Key Laboratory for Geo-Environmental Monitoring of Coastal Zone of the National Administration of Surveying, Mapping and Geoinformation, Shenzhen University, Shenzhen 518060, China.}
\thanks{B. Liu and G. Qiu are with the College of Information Engineering \& Guangdong Key Laboratory of Intelligent Information Processing, Shenzhen University, Shenzhen 518060, China.}
\thanks{Qing Li and G. Qiu are also with the School of Computer Science, University of Nottingham, Nottingham NG8 1BB, UK.}
\thanks{Corresponding author: G. Qiu (e-mail: qiu@szu.edu.cn).}
}

\maketitle

\begin{abstract}  
With the rapid growing of remotely sensed imagery data, there is a high demand for effective and efficient image retrieval tools to manage and exploit such data.
In this letter, we present a novel content-based remote sensing image retrieval method based on Triplet deep metric learning convolutional neural network (CNN). By constructing a Triplet network with metric learning objective function, we extract the representative features of the images in a semantic space in which images from the same class are close to each other while those from different classes are far apart. In such a semantic space, simple metric measures such as Euclidean distance can be used directly to compare the similarity of images and effectively retrieve images of the same class. We also investigate a supervised and an unsupervised learning methods for reducing the dimensionality of the learned semantic features. We present comprehensive experimental results on two publicly available remote sensing image retrieval datasets and show that our method significantly outperforms state-of-the-art.
\end{abstract}

\begin{IEEEkeywords}
Deep learning, metric learning, content-based image retrieval (CBIR), remote sensing image retrieval (RSIR).
\end{IEEEkeywords}

%
\IEEEpeerreviewmaketitle

\section{Introduction}
%
%
%
%
\IEEEPARstart{W}{ith} 
the development of remote sensing technologies, the ability of remote sensing image acquisition has been largely enhanced.
The quantity and quality of remote sensing images have increased dramatically.
Consequently, remote sensing image retrieval (RSIR) has received increasing interest in the remote sensing community.

Early remote sensing image retrieval systems search images by geographical location, acquisition time, or sensor type which are not directly related to the visual content of images.
Owing to the development of content-based image retrieval (CBIR), which employs the features extracted directly from the visual content of images for retrieval tasks, content-based remote sensing image retrieval has thus also witnessed great advance in recent years \cite{yang_geographic_2013,ozkan_performance_2014,napoletano_visual_2018}.

One of the major issues in CBIR is to find discriminative and robust features from images. Traditional methods rely on handcrafted features, the design of which requires sufficient expert knowledge and is time-consuming.
Handcrafted features are also widely exploited as remote sensing image representations in RSIR works \cite{yang_geographic_2013,ozkan_performance_2014}.
Popular handcrafted features include global features such as spectral (color), texture, and shape features, and aggregated local features such as bag of visual words (BoVW) \cite{yang_geographic_2013}, vector of locally aggregated descriptors (VLAD) \cite{ozkan_performance_2014}, and Fisher vector (FV) \cite{napoletano_visual_2018}.

The development of deep learning has advanced many areas including content-based image retrieval.
Deep convolutional neural network (CNN) features are highly abstractive and contain high-level semantic information, which have been shown to have superior performances to traditional handcrafted features in remote sensing image retrieval \cite{napoletano_visual_2018,ye_remote_2018,zhou_patternnet_2018}.
Furthermore, the deep features are learned automatically from data and there is no need for human effort in designing the features, which makes deep learning technique extremely valuable in large-scale remote sensing image retrieval.

Deep metric learning (DML) is an emerging technique that combines deep learning and metric learning \cite{schroff_facenet_2015}.
It exploits the discriminative power of deep neural networks to embed the images into an embedding metric space in which simple metrics such as the Euclidean distance can be used directly to measure the semantic similarity between images.
Deep metric learning is proven to be effective in fields like face recognition \cite{schroff_facenet_2015}, person re-identification \cite{hermans_defense_2017}, and natural image retrieval \cite{gordo_end-to-end_2017}.
Although remote sensing images are very different from ordinary natural images, DML still shows promising potential for content-based remote sensing image retrieval \cite{roy_deep_2018}.

In this letter, we present a novel Triplet deep neural network based metric learning method to enhance RSIR. Using deep convolutional neural networks, we embed the remote sensing images into a semantic space in which images from the same class are close to each other and those from different classes are far apart. We also investigate methods based on the use of a fully-connected layer of the CNN (supervised learning) and PCA (unsupervised learning) for reducing the dimensionality of the learned semantic features. We present comprehensive experiments on the popular UCMD \cite{yang_bag-of-visual-words_2010} and the large PatternNet \cite{zhou_patternnet_2018} RSIR datasets and the results demonstrate that our method significantly outperforms state-of-the-art.

\section{Methodology}
\label{sec:methodology}

In this section, we describe the details of the proposed Triplet network for RSIR, including the architecture and loss function of the network, as well as an effective and efficient method for selecting image triplets to train the network. We also present two methods, one based on a fully-connected layer of the CNN (supervised learning) and the other based on PCA (unsupervised learning), to reduce the dimension of extracted features from the proposed network for retrieval.

\subsection{Triplet Network for Image Retrieval}

\begin{figure}[h]
\centering
\includegraphics[width=0.46\textwidth]{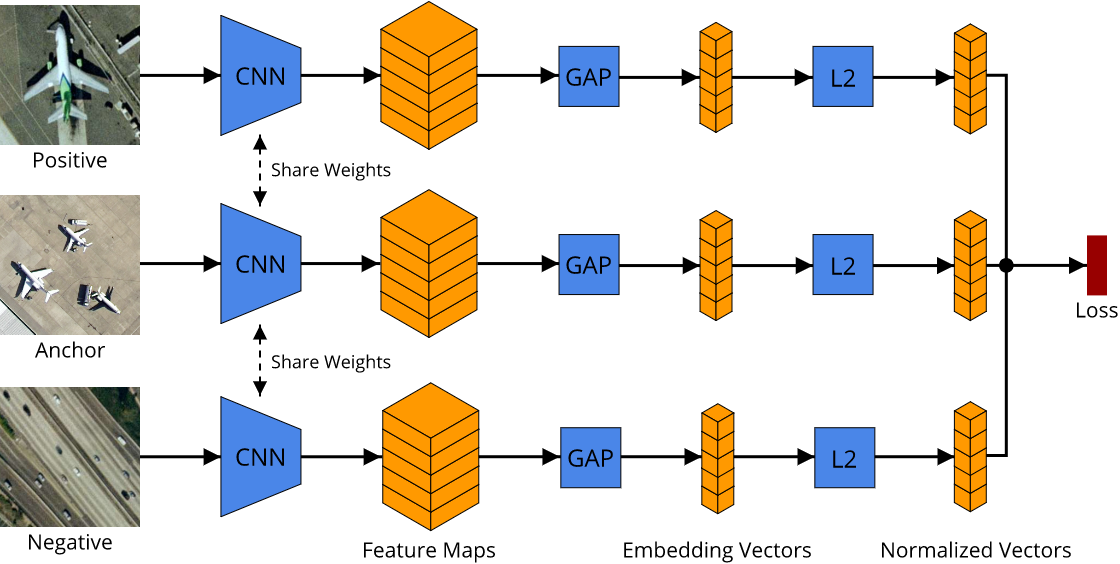}
\caption{The architecture of the proposed Triplet network. In the training phase, image triplets are fed into the three identical CNNs (with only convolutional layers) respectively to produce feature maps, which are then vectorized by global average pooling (GAP), and finally, the embedding vectors are $L_2$-normalized to compute the loss. In the testing phase, the $L_2$-normalized embedding vector can be exploited as the representative feature for retrieval.}
\label{fig:triplet-network-arch}
\end{figure}

\subsubsection{Network Architecture}

The overall architecture of our proposed Triplet network is shown in Fig. \ref{fig:triplet-network-arch}.
The network is composed of three identical convolutional neural networks which share the same weights. In the training phase, three sample images (a triplet) are fed into each network respectively, one image is set as \textit{anchor}, the other two are \textit{positive} and \textit{negative} samples. The \textit{positive} sample is of the same class as the \textit{anchor}, while the \textit{negative} sample belongs to a different class.
The three identical networks take input image triplets and output three corresponding feature maps. The feature maps are then vectorized by the global average pooling (GAP) operation to obtain fixed-length embedding features, which are further $L_2$-normalized.
Finally, the distance between the three extracted vectors will be used to compute the loss to train the network.
In the testing phase, images can be fed into one of the three identical networks to extract the fixed-length feature vectors for retrieval.

It should be noted that the CNNs used are ``fully convolutional'' which consist of only convolutional layers, so that the networks can extract features regardless of image size. This prevents the loss of information when resizing or cropping input images. Besides, ``fully convolutional'' CNNs are low-parameterized and fast to run.

\subsubsection{Loss Function}

The goal of the Triplet network is to learn a metric embedding function $f_{\theta}(x): \mathbb R^I \rightarrow \mathbb R^F$ that maps the input images to a feature space so that semantically similar images in $\mathbb R^I$ are metrically close in $\mathbb R^F$, where the function $f_{\theta}$ parameterized by $\theta$ represents the CNN-based feature extractor.
This is achieved by designing a triplet loss.

For an arbitrary image $I_i$, $x_i = f_{\theta}(I_i)$ is the corresponding feature vector in the metric embedding space.
Let the metric to measure the similarity of images in the embedding space be $d(x_i, x_j): \mathbb{R}^F \times \mathbb{R}^F \rightarrow \mathbb{R}$.
In this letter, squared Euclidean distance $d(x_i, x_j) = \left\| x_i - x_j \right\|_2^2$ is used as metric.
Then, for a triplet of images $T=(I_a, I_p, I_n)$, where $I_a$, $I_p$, $I_n$ are the anchor, positive, and negative images respectively, and $l_a, l_p, l_n$ ($l_a = l_p \neq l_n$) are their corresponding class labels, the corresponding embedding vectors are $(x_a, x_p, x_n) = (f_{\theta}(I_a), f_{\theta}(I_p), f_{\theta}(I_n))$, and thereby the triplet loss function can be formulated as follows:
\begin{equation}
\mathcal{L}(\theta) = \sum_{\substack{a,p,n \\ l_a = l_p \neq l_n}}{[d(x_a,x_p) - d(x_a,x_n) + m]_{+}},
\end{equation}
where $[x]_+$ represents $max(x, 0)$, $m$ denotes the margin. $d(x_a,x_p)$ and $d(x_a, x_n)$ are the distances between the anchor-positive and the anchor-negative pairs respectively, measured by the metric in the embedding space $\mathbb R^F$. 

\begin{figure}[h]
\centering
\includegraphics[width=0.4\textwidth]{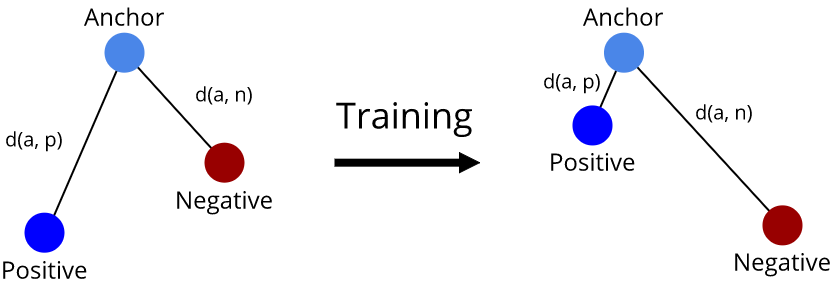}
\caption{Illustration of triplet loss. The training of network with triplet loss is the learning process of making the anchor and the positive sample (of the same class) closer while pushing away the negative sample (of different class).}
\label{fig:triplet-loss-demo}
\end{figure}

The intuition of the triplet loss is to make the positive sample closer to the anchor while push the negative sample far away. The process is illustrated in Fig. \ref{fig:triplet-loss-demo}.

\subsubsection{Triplet Selection for Effective Training}

It is crucial to select informative triplets for Triplet network training.
The naive way is to select triplets offline, by randomly picking positive samples from the same class as the anchor (except the anchor), and negative samples from any other classes.
However, this method is both inefficient and ineffective.
In a mini-batch with $3N$ images, there would only be $N$ triplets.

In this letter, we use the batch all triplet mining technique \cite{hermans_defense_2017} to select triplets. The key idea is to exhaust all the valid triplets within a mini-batch to compute the loss in the training phase.
In a mini-batch $\mathcal{I}$, let $N_C$ denotes the number of all the classes, $N_K$ denotes the number of images per class, then there would be $(N_K \times N_C) \times (N_K - 1) \times (N_K \times N_C - N_K)$ image triplets within the mini-batch. Let $I^i_a$ (anchor) and $I^i_p$ (positive) be the $a$-th and $p$-th image in class $i$, while $I^j_n$ (negative) be the $n$-th image in class $j$, then the loss function within the mini-batch can be formulated as follows:
\begin{equation}
\label{eq:loss-batch-all}
\mathcal{L}(\theta;\mathcal{I}) = \overbrace{\sum_{i=1}^{N_C} \sum_{a=1}^{N_K}}^{anchors}  \overbrace{\sum_{\substack{p=1 \\ p \neq a}}^{N_K}}^{pos.} 
\overbrace{\sum_{\substack{j=1 \\ j \neq i}}^{N_C} \sum_{n=1}^{N_K}}^{neg.} 
[d(x^i_a, x^i_p) - d(x^i_a, x^j_n) + m]_{+},
\end{equation}
where $(x^i_a,x^i_p,x^j_n) = (f_{\theta}(I^i_a),f_{\theta}(I^i_p),f_{\theta}(I^j_n))$ are the corresponding embedding vectors of image triplets $(I^i_a, I^i_p, I^j_n)$.

Compared with the naive offline triplet-selection strategy, the triplet mining method greatly increases the efficacy of training by exploiting all the valid triplets online within a training mini-batch. This makes full use of the hard samples of each training batch to compute loss, and thus makes the training process easier to converge. 

\subsection{Feature Dimension Reduction}

\begin{figure}[h]
\centering
\includegraphics[width=0.4\textwidth]{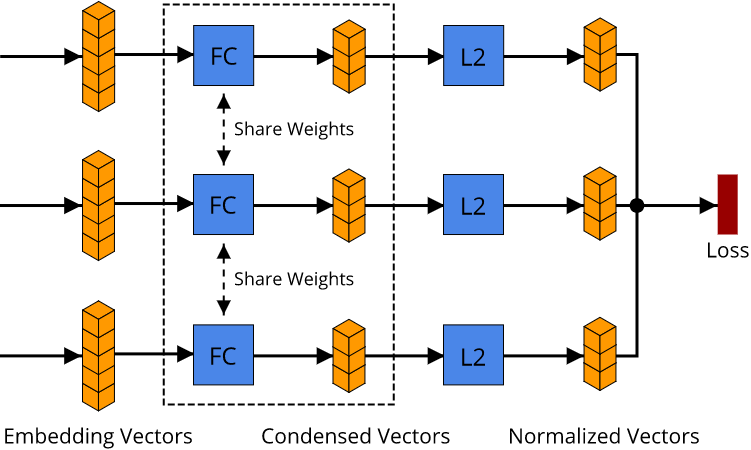}
\caption{Feature dimension reduction by adding fully-connected (FC) layers (enveloped by the dotted box) to the Triplet network (as shown in Fig. \ref{fig:triplet-network-arch}).}
\label{fig:fc-reduction}
\end{figure}

Deep features directly extracted from the proposed Triplet network are of high dimensions, which impacts the efficiency of similarity search in image retrieval and requires large storage.
To address this issue, we use a method based on the fully-connected (FC) layer of the CNN (supervised learning), and a method based on the traditional PCA (unsupervised learning), to reduce the dimensionality of the semantic features.

The process of employing a fully-connected layer of the CNN for dimensionality reduction is illustrated in Fig. \ref{fig:fc-reduction}.
In the training phase, the embedding vectors (extracted from the network in Fig. \ref{fig:triplet-network-arch}) are fed to a fully-connected layer of a lower dimension, which are then $L_2$-normalized to compute the final loss.
In the testing phase, the $L_2$-normalized condensed vector can be exploited as the representative feature for retrieval.

For PCA dimension reduction, the training images are firstly fed into the trained DML network to obtain the embedding feature vectors.
The covariance matrix of these vectors are computed and its eigenvectors are then used to project the feature vectors onto a lower dimensional space.
The features with dimension reduced by PCA are further $L_2$-normalized and then used for image retrieval.

\section{Experiments}
\label{sec:experiments}

\subsection{Datasets}

Two publicly available RSIR datasets, UCMD and PatternNet, are used to evaluate the proposed method.
The University of California, Merced (UCMD) land use dataset \cite{yang_bag-of-visual-words_2010} is the most widely used RSIR dataset. It includes 21 classes, with 100 images per class. All the images are 256$\times$256 pixels, and the pixel resolution is about 0.3$\mathrm{m}$. 
The images are extracted from the USGS Map from various US urban areas.
PatternNet \cite{zhou_patternnet_2018} is currently the largest publicly available dataset for remote sensing image retrieval. It consists of 38 classes, with 800 images per class, and the size of each image is 256$\times$256 pixels. The spatial resolution of the images ranges from 0.062 to 4.693 meters. The images are collected from Google Earth and Google Maps imagery from US cities.

\subsection{Experiment Setup}

For UCMD, we follow the data splitting that yields the best performance in \cite{ye_remote_2018}, which randomly selects 50\% images of each class for training and the rest 50\% for performance evaluation.
For PatternNet, we follow the 80\%/20\% training and testing data splitting strategy as per \cite{zhou_patternnet_2018}.

Three convolutional neural networks are employed as the basic networks for feature extraction, i.e. AlexNet \cite{krizhevsky_imagenet_2012}, VGG16 \cite{simonyan_very_2015}, and ResNet50 \cite{he_deep_2016}. For each network, only the convolutional layers are used to extract features, and global average pooling is operated on extracted output feature maps to obtain the final fixed-length feature vectors. The number of convolutional layers of the three CNNs and the dimension of their corresponding output features are listed in Table \ref{tab:net-config}.
\begin{table}[h!]
\centering
\caption{Configurations of CNN feature extractors}
\label{tab:net-config}
\begin{tabular}{lcc}
\hline
Network  & Conv. Layers & Feature Dimension \\ \hline
AlexNet  & 5            & 256       \\
VGG16    & 13           & 512       \\
ResNet50 & 49           & 2048      \\ \hline
\end{tabular}
\end{table}

In the experiments, the networks are implemented based on the PyTorch framework.
Adam optimizer is used to train the Triplet network, with a learning rate of 0.00001 and a batch size of 30. 
The parameters of the networks are initialized by the corresponding network weights pretrained on ImageNet \cite{deng_imagenet_2009}.
The maximum training iteration is set to 30 epochs.
The margin of the triplet loss is empirically set to 0.2 which was found to work consistently well in all experimental settings on both datasets.
Random horizontal and vertical image flips are applied as data augmentation. 

For comparison, the features extracted by the pretrained and the fine-tuned CNNs (with convolutional layers only) are used as baselines. For fine-tuning, fully-connected layers and partial convolutional layers (last layer for AlexNet, last two layers for VGG16, and last three layers for ResNet50) are retrained on the training sets of UCMD and PatternNet respectively.

\subsection{Performance Evaluation Metrics}

To evaluate image retrieval performance, average normalized modified retrieval rank (ANMRR) \cite{yang_geographic_2013}, mean average precision (mAP), and precision at $k$ ($P@k$, precision of the top-$k$ retrieval results) are utilized.
It should be noted that the lower the value of ANMRR is, the better the retrieval performance,
while it is opposite for mAP and $P@k$.

\subsection{Results and Analysis}

We evaluate our deep metric learning (DML)-based features by comparing results with baseline features from pretrained (PT) and fine-tuned (FT) CNNs, and results reported by previous works.

\subsubsection{Overall Results}

\begin{table}[h!]
\centering
\caption{Overall evaluation results on the UCMD dataset}
\label{tab:eval-ucmd}
\resizebox{0.48\textwidth}{!}{%
\begin{tabular}{lccccccc}
\hline
Features                            & ANMRR           & mAP             & P@5             & P@10            & P@50            & P@100           & P@1000          \\ \hline
BoVW \cite{yang_geographic_2013}    & 0.5910          &                 &                 &                 &                 &                 &                 \\
VLAD \cite{ozkan_performance_2014}  & 0.4604          &                 &                 &                 &                 &                 &                 \\
Fc7\_W (50)  \cite{ye_remote_2018}  & 0.0673          &                 &                 &                 &                 &                 &                 \\
Pool5\_W (50) \cite{ye_remote_2018} & 0.0404 	      &                 &                 &                 &                 &                 &                 \\ \hline
PT (AlexNet)   & 0.4534          & 0.4706          & 0.7787          & 0.6930          & 0.4379          & 0.2959          & 0.0490          \\
PT (VGG16)     & 0.4073          & 0.5179          & 0.8036          & 0.7283          & 0.4774          & 0.3204          & 0.0490          \\
PT (ResNet50)  & 0.3743          & 0.5532          & 0.8261          & 0.7620          & 0.5120          & 0.3348          & 0.0490          \\
FT (AlexNet)   & 0.2650          & 0.6751          & 0.8516          & 0.8147          & 0.6222          & 0.3875          & 0.0490          \\
FT (VGG16)     & 0.1478          & 0.8207          & 0.9076          & 0.8885          & 0.7705          & 0.4347          & 0.0490          \\
FT (ResNet50)  & 0.0612          & 0.9166          & 0.9589          & 0.9508          & 0.8663          & 0.4724          & 0.0490          \\
DML (AlexNet)  & 0.1651          & 0.7792          & 0.8903          & 0.8670          & 0.7213          & 0.4388          & 0.0490          \\
DML (VGG16)    & 0.0341          & 0.9487          & 0.9741          & 0.9687          & 0.9057          & 0.4828          & 0.0490          \\
DML (ResNet50) & \textbf{0.0223} & \textbf{0.9663} & \textbf{0.9775} & \textbf{0.9757} & \textbf{0.9320} & \textbf{0.4855} & \textbf{0.0490} \\ \hline
\end{tabular}}
\end{table}

\begin{table}[h!]
\centering
\caption{Overall evaluation results on the PatternNet dataset}
\label{tab:eval-patternnet}
\resizebox{0.48\textwidth}{!}{%
\begin{tabular}{lccccccc}
\hline
Features      				  & ANMRR           & mAP             & P@5             & P@10            & P@50            & P@100           & P@1000          \\ \hline
Gabor Texture \cite{zhou_patternnet_2018} & 0.6439          & 0.2773          & 0.6855          & 0.6278          & 0.4461          & 0.3552          & 0.0899          \\
VLAD          \cite{zhou_patternnet_2018} & 0.5677          & 0.3410          & 0.5825          & 0.5570          & 0.4757          & 0.4111          & 0.1104          \\
UFL           \cite{zhou_patternnet_2018} & 0.6584          & 0.2535          & 0.5209          & 0.4882          & 0.3811          & 0.3192          & 0.0979          \\
VGGF\_Fc1     \cite{zhou_patternnet_2018} & 0.3177          & 0.6195          & 0.9246          & 0.9037          & 0.7926          & 0.6905          & 0.1425          \\
VGGF\_Fc2     \cite{zhou_patternnet_2018} & 0.2995          & 0.6337          & 0.9152          & 0.8964          & 0.7999          & 0.7047          & 0.1452          \\
VGGS\_Fc1     \cite{zhou_patternnet_2018} & 0.3050          & 0.6328          & 0.9274          & 0.9070          & 0.8003          & 0.7013          & 0.1436          \\
VGGS\_Fc2     \cite{zhou_patternnet_2018} & 0.2961          & 0.6374          & 0.9192          & 0.9009          & 0.8021          & 0.7073          & 0.1455          \\
ResNet50      \cite{zhou_patternnet_2018} & 0.2584          & 0.6823          & 0.9413          & 0.9241          & 0.8371          & 0.7493          & 0.1464          \\
LDCNN         \cite{zhou_patternnet_2018} & 0.2416          & 0.6917          & 0.6681          & 0.6611          & 0.6747          & 0.6880          & 0.1408          \\ \hline
PT (AlexNet)   & 0.3352          & 0.5980          & 0.9226          & 0.9001          & 0.7872          & 0.6784          & 0.1367          \\
PT (VGG16)     & 0.3121          & 0.6217          & 0.9192          & 0.8993          & 0.7956          & 0.6991          & 0.1404          \\
PT (ResNet50)  & 0.2848          & 0.6512          & 0.9304          & 0.9125          & 0.8193          & 0.7266          & 0.1426          \\
FT (AlexNet)   & 0.0906          & 0.8842          & 0.9671          & 0.9625          & 0.9403          & 0.9125          & 0.1564          \\
FT (VGG16)     & 0.0521          & 0.9328          & 0.9766          & 0.9729          & 0.9595          & 0.9464          & 0.1575          \\
FT (ResNet50)  & 0.0258          & 0.9651          & 0.9893          & 0.9879          & 0.9823          & 0.9751          & 0.1584          \\
DML (AlexNet)  & 0.0129          & 0.9811          & 0.9882          & 0.9874          & 0.9852          & 0.9835          & 0.1590          \\
DML (VGG16)    & 0.0036          & 0.9943          & 0.9953          & 0.9950          & 0.9947          & 0.9946          & 0.1590          \\
DML (ResNet50) & \textbf{0.0030} & \textbf{0.9955} & \textbf{0.9958} & \textbf{0.9957} & \textbf{0.9957} & \textbf{0.9954} & \textbf{0.1590} \\ \hline
\end{tabular}}
\end{table}

The overall results on the UCMD and PatternNet datasets are shown in Table \ref{tab:eval-ucmd} and Table \ref{tab:eval-patternnet} respectively.
As can be seen, in general, DML-based features achieve the best performance, fine-tuned CNN features achieve competitive results, while pretrained features show the worst results. In addition, the deeper the network is, the better the image retrieval performance.
It can be seen that DML-based ResNet50 features have achieved the best results on both datasets, significantly outperformed all the other methods on all the evaluation metrics.

It is interesting to note that the best performance on PatternNet is significantly better than that on the UCMD dataset. One probable reason is that deep metric learning is data hungry and the amount of training data influences the learning of representative features. Since PatternNet is much larger than UCMD, the network for the PatternNet dataset is better trained than that for the UCMD dataset.

\subsubsection{Per-class Results}

\begin{figure}[h!]
\centering
\includegraphics[width=0.48\textwidth]{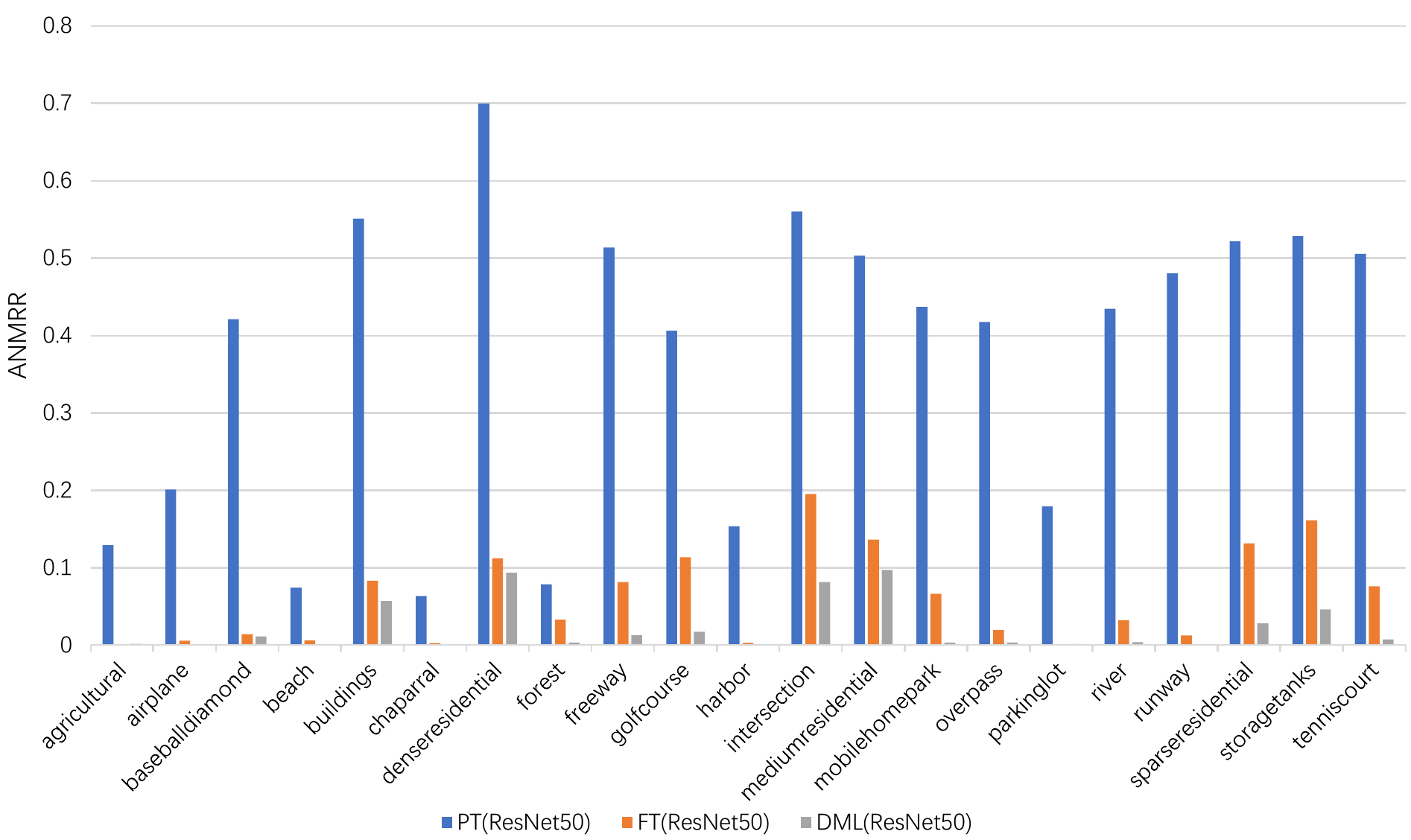}
\caption{The per-class results of different features on the UCMD dataset.}
\label{fig:eval-class-ucmd}
\end{figure}

\begin{figure}[h!]
\centering
\includegraphics[width=0.48\textwidth]{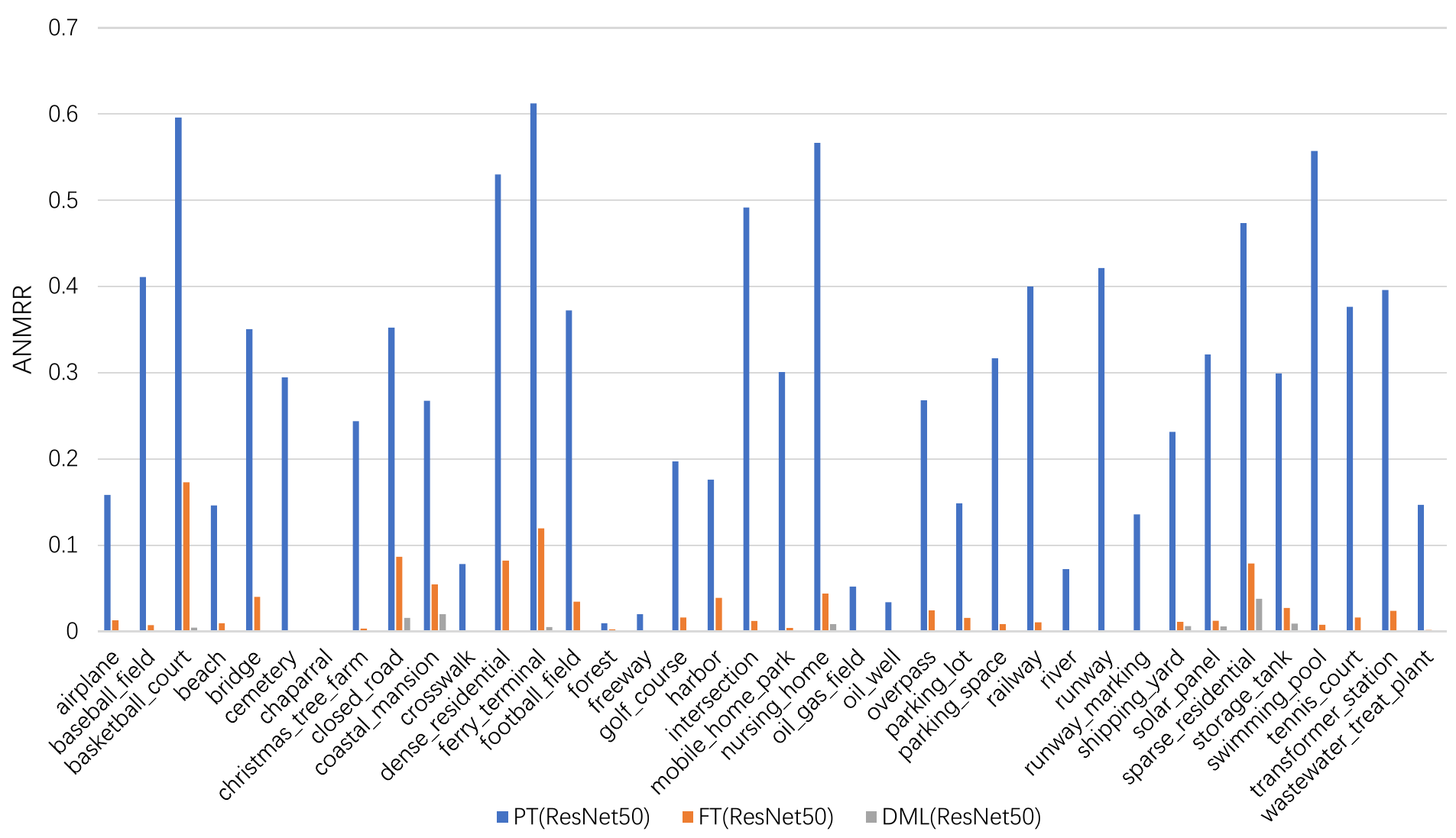}
\caption{The per-class results of different features on the PatternNet dataset.}
\label{fig:eval-class-patternnet}
\end{figure}

The per-class performances in terms of ANMRR of different ResNet50-based deep features on the two datasets are shown in Fig. \ref{fig:eval-class-ucmd} and Fig. \ref{fig:eval-class-patternnet} respectively.
As presented in Fig. \ref{fig:eval-class-ucmd}, in general, for almost every class, DML-based features outperform fine-tuned features, and both of them perform much better than pretrained features.
Pretrained ResNet50-based features have particular difficulty in retrieving images of \textit{buildings}, \textit{dense residential}, and \textit{intersection} classes, with an average ANMRR of 0.60, much higher than that of its counterpart, with 0.13 for the fine-tuned features, and less than 0.08 for DML-based features.
It can be seen from Fig. \ref{fig:eval-class-patternnet} that DML-based features outperform fine-tuned features significantly, while the latter perform much better than pretrained features for all the classes.
Pretrained ResNet50-based features perform poorly on classes like \textit{basketball court}, \textit{ferry terminal}, and \textit{nursing home}, with an average ANMRR of 0.59. This value for the fine-tuned features is 0.11, while it is less than 0.006 for DML-based features, which further demonstrates the superior performance of DML-based features for content-based remote sensing image retrieval.

\subsubsection{Qualitative retrieval results}

\begin{figure}[h]
\centering
\includegraphics[width=0.48\textwidth]{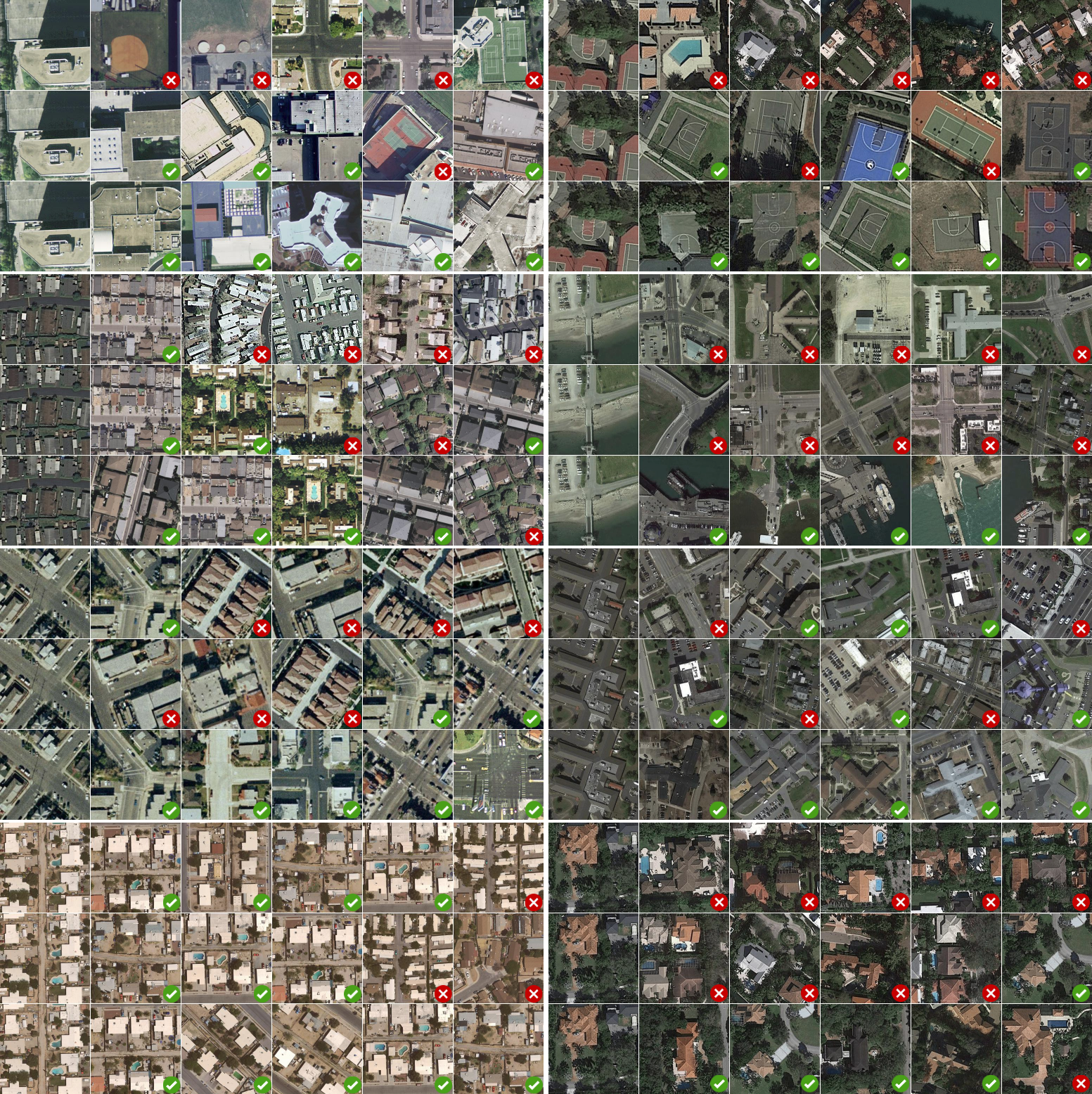}
\caption{Top-5 retrieval results for the UCMD (left column) and PatternNet (right column) datasets. For both columns, the first image is the query image, and for each query, the three rows are results for the extracted features from pretrained, fine-tuned, and DML-based ResNet50 respectively. The small green ticks and red crosses indicate correct and false results respectively.}
\label{fig:case-study}
\end{figure}

As can be seen in Fig. \ref{fig:eval-class-ucmd} and Fig. \ref{fig:eval-class-patternnet}, DML-based features perform significantly better than pretrained and fine-tuned CNN features for almost all the classes on both datasets.
To investigate into the details, qualitative results of several difficult query cases are presented in Fig. \ref{fig:case-study}, which shows the top-5 retrieved images that are similar to the query image, using features extracted by pretrained, fine-tuned, and DML-based ResNet50 networks respectively.
In Fig. \ref{fig:case-study}, the left column is the results on the UCMD dataset, and query images are from the classes of \textit{buildings}, \textit{dense residential}, \textit{intersection}, and \textit{medium residential}; the query results of the PatternNet dataset are presented in the right column, with query images from the classes of \textit{basketball court}, \textit{ferry terminal}, \textit{nursing home}, and \textit{sparse residential}.
DML-based features outperform pretrained and fine-tuned features on the cases noticeably.
The results indicate that deep metric learning can well learn high-level semantic features to distinguish images with high within-class variance.

\subsection{Analysis of Feature Dimension Reduction}

\begin{figure}[h]
\centering
\includegraphics[width=0.48\textwidth]{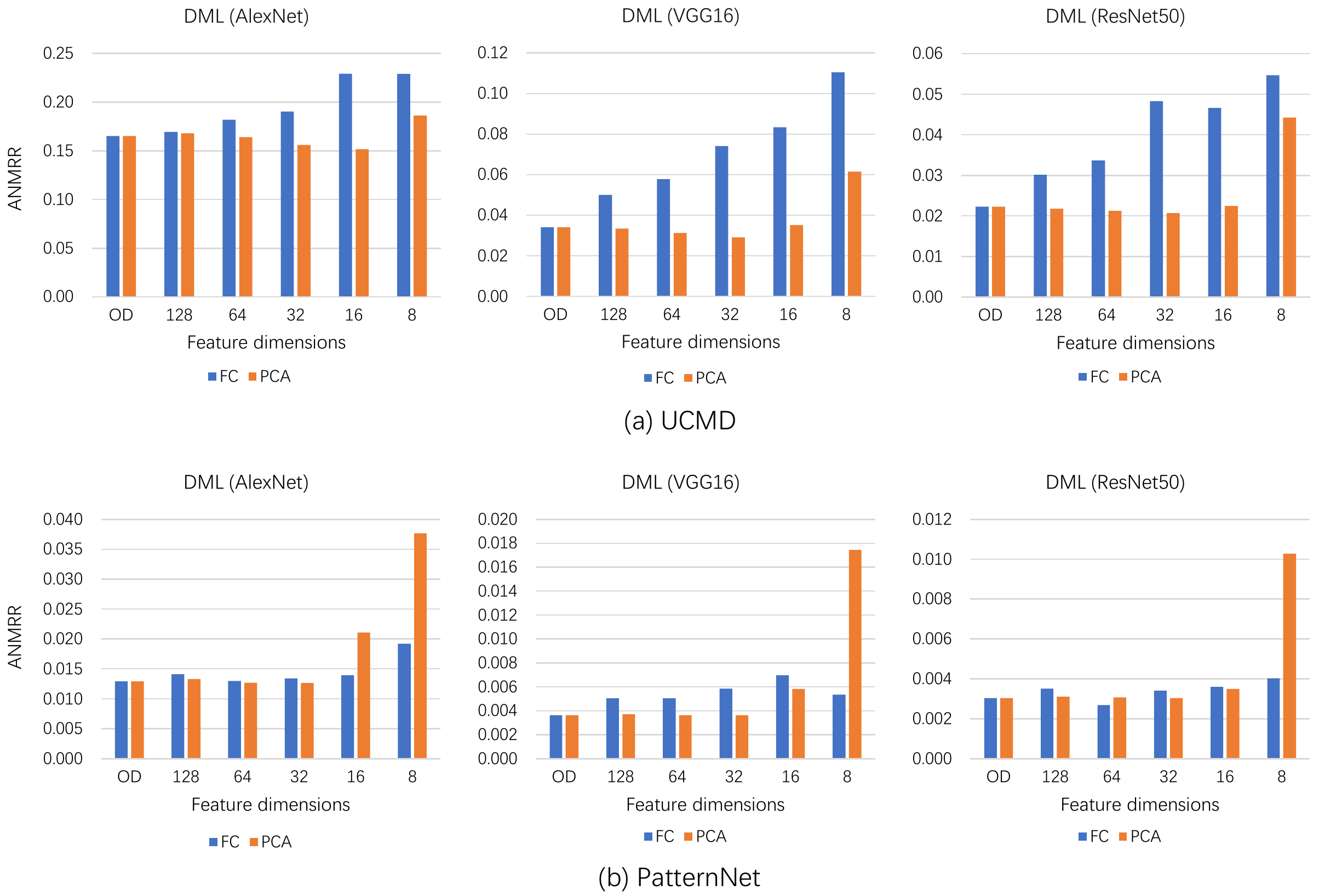}
\caption{Performance of deep metric learning-based features with different dimensions on (a) UCMD and (b) PatternNet. (OD: original dimension)}
\label{fig:feat-dim}
\end{figure}

Fully-connected (FC) layers and principal component analysis (PCA) are used to reduce the dimension of extracted features.
The retrieval performances of DML-based features with different dimensions (i.e. the original dimension, the reduced dimensions of 128, 64, 32, 16, and 8) using the two methods are shown in Fig. \ref{fig:feat-dim}.
As can be seen, for PCA, the reduction of feature size from the original dimension to 32 has relatively small impact on retrieval performances on both UCMD and PatternNet, the best performances are generally achieved at 32, and the performances drop significantly when the dimension is further reduced since the values of ANMRR increase noticeably.
For FC-based dimension reduction, the ANMRR values generally increase with the reduction of feature dimension on the UCMD dataset, and the performance is worse than PCA at the same dimension. However, FC-based reduction performances are generally stable across different dimensions on the PatternNet dataset, and the performance is much better than PCA when feature dimension is reduced to less than 16, which implies that FC-based feature dimension reduction performs better with sufficient training data.

\section{Conclusions}
\label{sec:conclusions}

Content-based remote sensing image retrieval is key to effective use of the ever-growing remote sensing images.
In this letter, we use deep metric learning-based Triplet network to learn deep metric embeddings from positive and negative sample images, and thereby enhance the retrieval performance of remote sensing images.
We test our methods on two publicly available datasets and achieve state-of-the-art performances on both datasets.
We also investigate supervised CNN fully-connected layers and unsupervised PCA methods to further reduce the dimension of extracted features.
We have successfully demonstrated the effectiveness of deep metric learning method for remote sensing image retrieval.




\ifCLASSOPTIONcaptionsoff
  \newpage
\fi



%

\end{document}